\documentclass[runningheads]{llncs}

\usepackage{eccv}

\usepackage{eccvabbrv}

\usepackage{pifont} 
\usepackage{graphicx}
\usepackage{booktabs}
\usepackage{colortbl}
\usepackage{multirow}
\usepackage{enumitem}
\usepackage[accsupp]{axessibility}  

\usepackage[pagebackref,breaklinks,colorlinks,citecolor=eccvblue]{hyperref}
\newcommand{\ModelName}{{F-HOI}\xspace}
\newcommand{\DataName}{{Semantic-HOI}\xspace}
\usepackage{orcidlink}
\begin{document}

\title{
F-HOI: Toward Fine-grained Semantic-Aligned 3D Human-Object Interactions
} 

\titlerunning{F-HOI}


\author{Jie Yang\textsuperscript{1,2,$\star$}\orcidlink{0009-0009-5891-2911}\and%
Xuesong Niu\textsuperscript{2,$\star$}\orcidlink{0000-0001-7737-4287} \and %
Nan Jiang\textsuperscript{2,3,$\star$}\orcidlink{0009-0006-5726-7672} \and \\
Ruimao Zhang\textsuperscript{1,\textdagger}\orcidlink{0000−0001−9511−7532} %
\and
Siyuan Huang\textsuperscript{2,\textdagger}\orcidlink{0000-0003-1524-7148}}

\authorrunning{J.~Yang, et al.}

\institute{\textsuperscript{1}The Chinese University of Hong Kong, Shenzhen~~~\textsuperscript{2}State Key Laboratory of General Artificial Intelligence, BIGAI~~~\textsuperscript{3}Institute for AI, Peking University 
\url{https://f-hoi.github.io}}

\def\thefootnote{}\footnotetext{\textsuperscript{$\star$}Equal contribution~\textsuperscript{\textdagger}Corresponding author}


\maketitle

\begin{abstract}
Existing 3D human object interaction (HOI) datasets and models simply align global descriptions with the long HOI sequence,
while lacking a detailed understanding of intermediate states and the transitions between states. 
In this paper, we argue that fine-grained semantic alignment, which utilizes state-level descriptions, offers a promising paradigm for learning semantically rich HOI representations.
To achieve this, we introduce Semantic-HOI, a new dataset comprising over 20K paired HOI states with fine-grained descriptions for each HOI state and the body movements that happen between two consecutive states. 
Leveraging the proposed dataset, we design three state-level HOI tasks to accomplish fine-grained semantic alignment within the HOI sequence.
Additionally, we propose a unified model called \ModelName, designed to leverage multimodal instructions and empower the Multi-modal Large Language Model to efficiently handle diverse HOI tasks. 
\ModelName offers multiple advantages: (1) It employs a unified task formulation that supports the use of versatile multimodal inputs. (2) It maintains consistency in HOI across 2D, 3D, and linguistic spaces. (3) It utilizes fine-grained textual supervision for direct optimization, avoiding intricate modeling of HOI states. Extensive experiments reveal that \ModelName effectively aligns HOI states with fine-grained semantic descriptions, adeptly tackling understanding, reasoning, generation, and reconstruction tasks.

\end{abstract}

\section{Introduction}
\label{intro}
\begin{figure}[h]
    \centering
    \includegraphics[width=\linewidth]{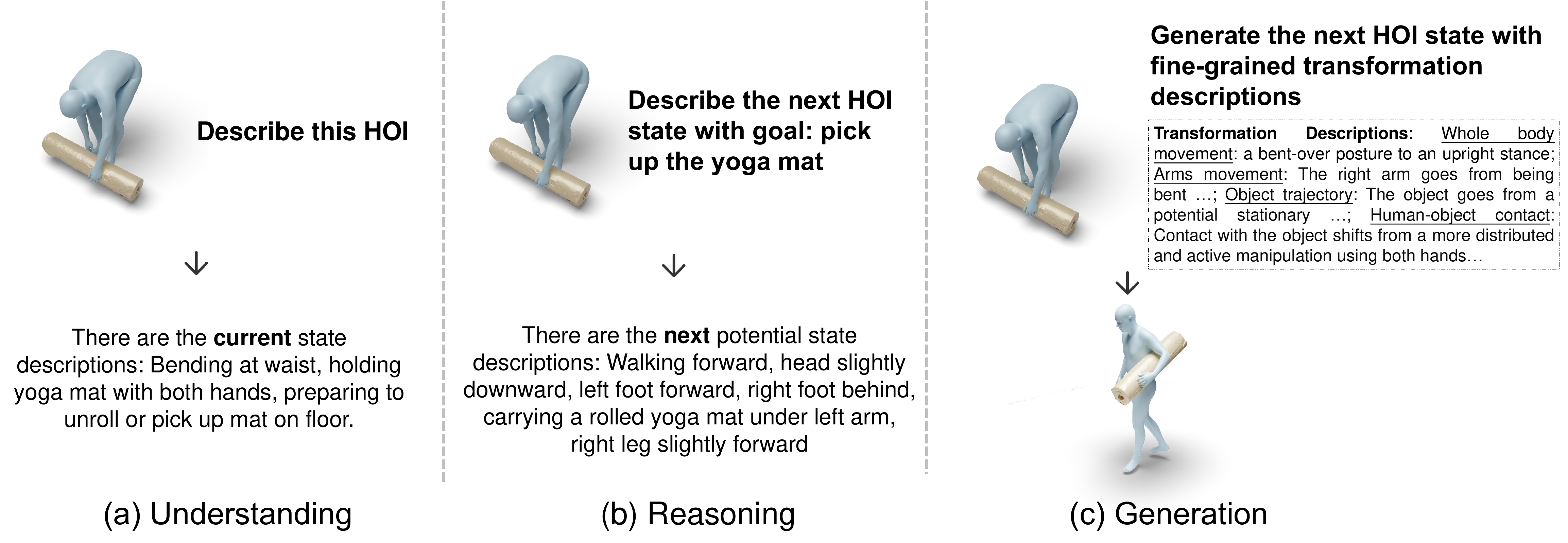}
        \vspace{-0.5cm}
    \caption{Illustration of three state-level tasks to achieve fine-grained semantic alignment.
    }
    \label{fig:fig1}
    \vspace{-0.5cm}
\end{figure}

Modeling human-object interaction (HOI) in 3D space 
is critical for various downstream applications, such as computer animation, virtual reality, and embodied AI~\cite{stacey2012animation, geijtenbeek2012interactive,mourot2022survey, wang2023physhoi,yang2023semantic}. 
The HOI process involves a sequence of continuous state transitions, which contains intricate changes in body parts, object trajectories, and interaction contacts. 
However, existing models~\cite{peng2023hoi,bhatnagar2022behave,wang2022humanise,taheri2020grab} only align global descriptions with such processes, making them struggle to comprehend each HOI state and the transitions between states in the fine-grained semantic space. In the literature, how to achieve fine-grained semantic-aligned 3D HOI is still a challenging yet under-explored issue due to the following difficulties:
\begin{enumerate}
\item 
\textbf{Dataset Gap.} Existing HOI datasets only provide coarse-grained goal descriptions (e.g., ``a person picks up a backpack'') to depict a long HOI sequence. Thus, the absence of fine-grained semantic descriptions significantly hampers progress in the related field.
\item
\textbf{Model Capacity.} Aligning the fine-grained semantic descriptions with HOIs is non-trivial. It requires a model that can establish alignments from a limited dataset, possesses powerful semantic comprehension skills to handle extensive textual descriptions, and has prior knowledge of real-world actions.
\end{enumerate}

To address the first issue, we reconstruct a new dataset named \DataName from three existing datasets, bridging the semantic gap in current datasets by furnishing fine-grained descriptions for HOI states and detailing the movements between two consecutive HOI states. Based on the proposed dataset, we design three state-level tasks to achieve fine-grained HOI modeling from different perspectives:  \textbf{(1) Understanding:} None of the current tasks explicitly involve understanding an HOI state (\textit{i.e.}, human pose with object pose) via textual descriptions. Thus, we are motivated to propose such a task and aim to achieve fine-grained understanding, as shown in \cref{fig:fig1}-(a).
\textbf{(2) Reasoning:} Building upon the understanding task, we further increase the level of difficulty by describing the next HOI state given the current HOI state and the overall HOI goal, as shown in \cref{fig:fig1}-(b).
 \textbf{(3) Generation:} Beyond the understanding and reasoning tasks, we further explore the fine-grained action control, which aims to leverage the transformation descriptions to generate the next state from the current one, as shown in \cref{fig:fig1}-(c).

To address the second problem,  we introduce \ModelName, a novel unified framework to tackle diverse 3D HOI tasks. Specifically, \ModelName first integrates various input modalities, including 2D images, 3D object meshes, 3D HOI-Pose (comprising human and object poses), and textual descriptions, into a unified architecture.
By employing different task instructions in the training phase, it progressively learns consistent HOI representations across 2D, 3D, and linguistic spaces, and realizes the mutual enhancement of different tasks.  
Once the model is optimized, it can leverage the powerful language understanding capabilities inherent in the multi-modal large language model to adeptly execute diverse HOI tasks with flexible inputs.

Through extensive experiments, we demonstrate that F-HOI can effectively align HOI states of sequences with fine-grained semantic descriptions, adeptly tackling understanding, reasoning, and generation tasks, along with the traditional reconstruction task. In addition, our ablation studies reveal that our model designs, coupled with the proposed dataset and training strategies, could improve fine-grained 3D HOI modeling. Finally, as a pioneering work, we provide a comprehensive discussion to inspire future research in the related field.

In summary, the contributions of this work are three-fold:
\begin{itemize}
\item As far as we know, this is the first work to explore the problem of fine-grained semantic-aligned 3D HOI modeling. To tackle such a problem, we introduce a new dataset named Semantic-HOI to bridge the annotation gap present in current datasets by providing fine-grained descriptions for HOI states and the body movement between two consecutive states.

\item To learn and evaluate the fine-grained HOI representation, we define three new state-level HOI tasks from the perspectives of understanding, reasoning, and generation. Furthermore, we present F-HOI, which empowers the MLLM to execute the above HOI tasks with flexible inputs.

\item Extensive experiments show \ModelName can effectively align HOI states with fine-grained semantic descriptions. We hope our proposed dataset and tasks can bring in
new perspectives to fine-grained semantic-aligned HOI modeling.

\end{itemize}

\section{Related Work}
\subsection{Human-Object Interaction}
Research in Human-Object Interaction (HOI) has traditionally focused on identifying interactions from images~\cite{chao2015hico,gkioxari2018detecting,liao2022gen,yuan2022detecting,huang2023diffusion,yang2024open,yang2023boosting}, 3D interaction reconstruction~\cite{wang2022reconstructing,savva2016pigraphs,hassan2019resolving,chen2019holistic,weng2021holistic,xu2021d3d,zanfir2018monocular,siwei2021learning} and generation~\cite{hassan2021populating,wang2021synthesizing,xu2020hierarchical,holden2017phase,wang2022humanise,zhang2022couch,diller2023cg,xiao2023unified,jiang2024scaling}. Some noteworthy contributions include Phosa~\cite{zhang2020perceiving}, which geometrically reconstructs HOIs by utilizing contact priors from different body regions, and GOAL~\cite{taheri2022goal}, which employs a conditional variational autoencoder (cVAE) to generate full-body motions for object grasping by estimating the grasping pose for the entire body. CHOIS~\cite{li2023controllable} further extends the field by synthesizing HOI motions using a conditional diffusion model based on language descriptions and the initial state of the object and the human involved. 
 On the other hand, the advent of 3D HOI assets~\cite{zhang2020perceiving} and datasets, which include visual recordings~\cite{jiang2023full,xu2021d3d,hassan2019resolving,jiang2024scaling}, text annotations~\cite{li2023object,taheri2020grab} or both~\cite{bhatnagar2022behave,wang2022humanise}, have promoted effective HOI modeling across various applications. However, existing models and datasets typically rely on coarse-grained descriptions for interactions, making it challenging to learn fine-grained semantic alignment. Our work represents the \textbf{first} attempt to address this limitation by constructing a new dataset with rich descriptions for body parts, objects, and their interactions, and proposing three new fine-grained state-level HOI tasks from different perspectives.

\subsection{Multimodal Large Language Models}
Large Language Models (LLMs) are rapidly emerging as powerful tools across various domains. While leading models like OpenAI's ChatGPT~\cite{chatgpt} and GPT-4~\cite{gpt4} remain proprietary, the availability of open-source LLMs such as Vicuna~\cite{vicuna}, LLaMA~\cite{touvron2023llama}, and Alpaca~\cite{alpaca} is enabling researchers to engage in multimodal research. In general, there are two technical paradigms to leverage the LLMs to solve multi-modality tasks: Firstly, LLMs can serve as effective decision-making agents by interfacing with task-specific models through API calls~\cite{yang2023mm,shen2023hugginggpt,liu2023internchat,yang2023gpt4tools,wang2023visionllm,detgpt,2023audiogpt}. Through carefully designed prompt engineering or instruction tuning, LLMs can generate API calls to address multi-modal tasks. However, this approach may not fully comprehend the intricacies of task-specific modalities, leading to potential failures when dealing with complex scenes. {Secondly,} an advanced approach is to map modality-specific representations into the language embedding space of the LLM. Recent works like LLaVA~\cite{llava,liu2023improvedllava} and MiniGPT-4~\cite{zhu2023minigpt} incorporate pre-trained visual encoders to obtain image features and train projection layers to align visual representations with the language space of LLM. This approach can also be extended to speech generation~\cite{2023speechgpt}, image generation~\cite{zheng2023minigpt,huang2023smartedit}, video understanding~\cite{2023videochat, 2023videollama}, and other perception tasks~\cite{lai2023lisa,zhang2023llava,xu2023pixel,pi2023perceptiongpt,feng2023posegpt}, providing a more comprehensive understanding of multi-modal data. Among these, ChatPose~\cite{feng2024chatpose} is most relevant to our work, as it explores the use of 3D body pose as a new modality for LLMs to process. However, our work goes further by comprehensively considering humans, objects, and their interactions. More importantly, we emphasize exploring the problem of fine-grained semantic-aligned HOI modeling. To achieve this, we leverage multi-modal instructions to empower MLLMs to complete fine-grained HOI tasks, thereby demonstrating significant potential.

\section{Dataset}
\label{dataset_all}
\subsection{Motivation}
\label{dataset_motivate}

\begin{table*}[t!]
    \centering
    \small
            \caption{Statistics of \DataName collected from three existing datasets.}
            \vspace{-0.3cm}
    \setlength{\tabcolsep}{3pt}
    \resizebox{\linewidth}{!}{%
        \begin{tabular}{l|c|cc|cc|c}
            \toprule
            
            \multirow{2}{*}{Datasets} & \multirow{2}{*}{2D Image} & \multicolumn{2}{c|}{3D HOI} & \multicolumn{2}{c|}{Text Descriptions} & \multirow{2}{*}{\# Unified HOI Pairs} \\
            \cmidrule(lr){3-4} \cmidrule(lr){5-6}
            & & \# 3D Object & 3D Human Pose  & Goal Descriptions & Fine-grained Descriptions &\\
            \midrule
            GRAB~\cite{taheri2020grab}  & \ding{55} & \checkmark    & \checkmark  & \checkmark & \ding{55} & 1187 \\
            CHAIRS~\cite{jiang2023full} & \checkmark & \checkmark & \checkmark  &  \checkmark & \ding{55} &  3368 \\ 
            BEHAVE~\cite{bhatnagar2022behave} & \checkmark & \checkmark   & \checkmark   & \checkmark & \ding{55} & 15886 \\
            \hline
                        \DataName & \checkmark & \checkmark  &\checkmark  &  \checkmark &  \checkmark &  20441 \\
            \bottomrule
        \end{tabular}%
    }%
        \vspace{-0.6cm}
    \label{tab:dataset}
\end{table*}

As illustrated in \cref{tab:dataset}, existing datasets focus on different subsets of objects and interactions, while providing only goal descriptions for long-term interaction processes. 
These limitations restrict the effectiveness of models in handling diverse scenes and achieving fine-grained semantic alignments for HOIs.
To address these dataset gaps, we introduce a novel dataset called \DataName, characterized by two primary features:
(1) \textbf{Diverse objects}. We aggregate and unify data from three established datasets to incorporate a wider range of objects;
(2) \textbf{Detailed descriptions}. Instead of directly describing long-term HOI sequences, which demand extensive language to capture various redundant action changes, we provide detailed descriptions for each HOI state and highlight the body movements between consecutive HOI states.

\begin{figure}[t]
    \centering
    \includegraphics[width=\linewidth]{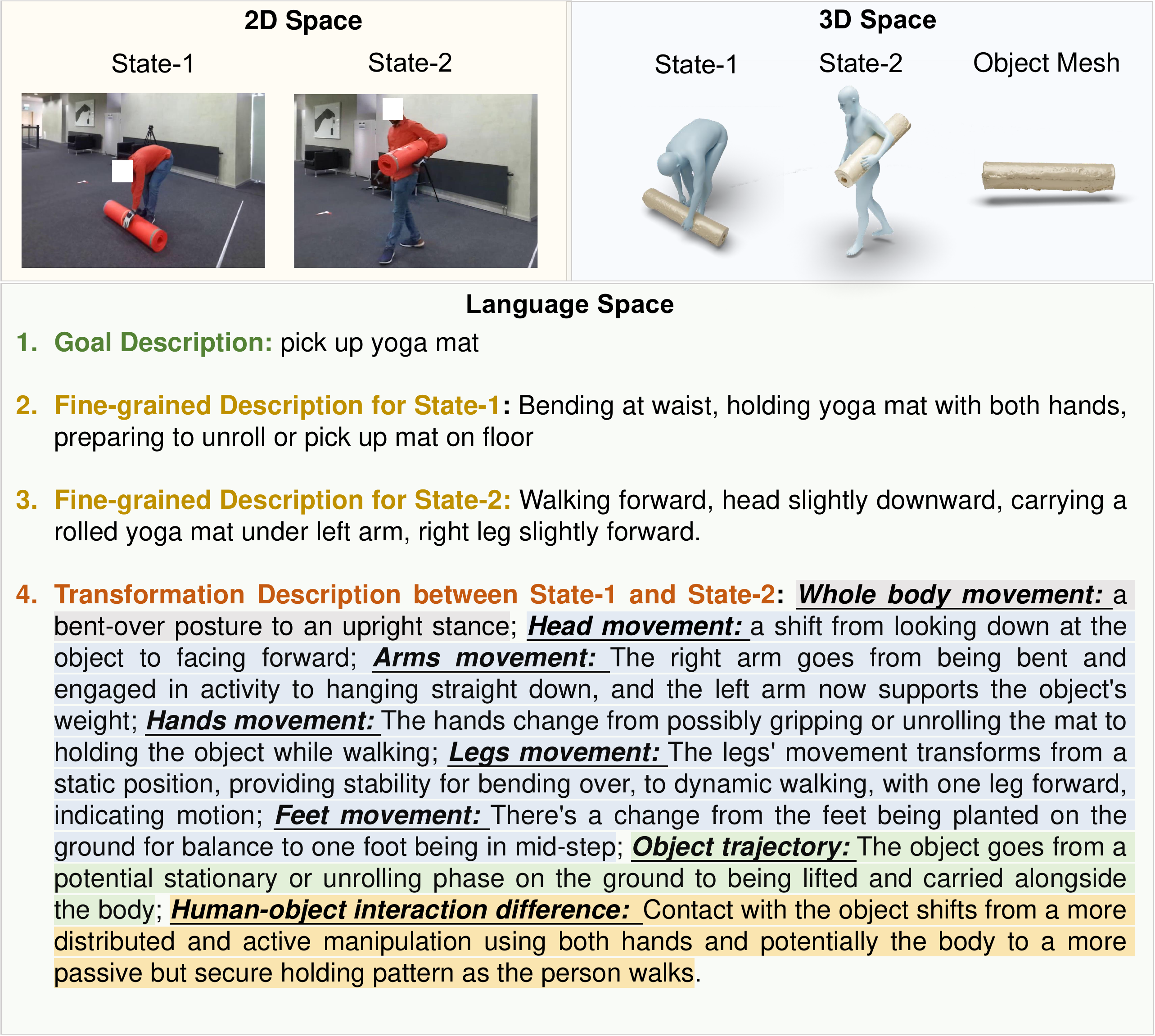}
    \vspace{-0.7cm}
    \caption{Illustration of the annotations in \DataName for a paired HOI sample.}
        \label{fig:annotations}
                \vspace{-0.5cm}
\end{figure}
\subsection{Data Collection}
We collect data from $3$ existing datasets (GRAB~\cite{taheri2020grab}, CHAIRS~\cite{jiang2023full} and BEHAVE~\cite{bhatnagar2022behave}), while carefully considering the following data balancing principles:
 
\noindent \textbf{Balanced Data for Interaction Diversity}: 
Each dataset in \cref{tab:dataset} is video-based and each video contains a single interaction process. To ensure diversity in interaction, we randomly sample a subset of state pairs from these videos.

\noindent \textbf{Balance Data for Images Discrepancies and Object Categories}: Since the GRAB dataset lacks natural images, we need to render 3D HOIs into 2D images. Additionally, the CHAIRS dataset predominantly features interactions with a diverse range of chairs. To address the variations in data distribution between rendered and natural images, and to ensure the super-category diversity of the interactive objects, we have restricted the number of samples drawn from both the GRAB and CHAIRS datasets to maintain balance. In contrast, we have included an increased number of samples from the BEHAVE dataset.

\subsection{Dataset Construction}
\noindent \textbf{Annotations.} As depicted in \cref{fig:annotations}, the annotations of an HOI pair comprises the following components:
\begin{enumerate}
\item 2D images for the current state and the next state.
\item HOI poses for the current state and the next state, along with object mesh.
\item Goal description for the action.
\item Fine-grained descriptions for both the current state and the next state.
\item Transformation descriptions detailing the changes between the current state and the next state.
\end{enumerate}
where components 1-3 can be derived from the original datasets. For components 4-5, we prompt GPT-4V~\cite{gpt4} using the given 2D images for annotations. During this process, we meticulously design the formats for fine-grained descriptions as follows: (a) decoupled human pose descriptions, including whole-body, head, two arms, two hands, two legs, and two feet; (b) object state descriptions; (c) interaction state descriptions.
Based on the above prompts, we can offer both part-level state descriptions and part-level movement descriptions. We also provide action descriptions to supplement ambiguous and incomplete goal descriptions.

\noindent \textbf{Statistics Analysis.}
Considering the potential for response errors in GPT-4V, we conduct manual verification to filter out improperly formatted fine-grained descriptions.
In total, our \DataName comprises $20,441$ pairs, with $1,187$ from GRAB, $3,368$ from CHAIRS, and $15,886$ from BEHAVE.
Furthermore, following BEHAVE, we split \DataName into about 70\%  for training and 30\% for testing to show the potential of fine-grained semantic-aligned HOI.

\begin{table}[t]
    \centering
        \caption{Comparison with related works about linking textual descriptions and human pose. Our dataset and annotation strategy are the first to explore the problem of fine-grained semantic-aligned 3D human-object interaction modeling.}
        \vspace{-0.3cm}
    \resizebox{0.8\linewidth}{!}{
        \setlength{\tabcolsep}{1mm}
        \begin{tabular}{l|cc|cccc}
            \toprule
            Method & Human & Object & Contact & State & Transition & Fine-grained \\
            \midrule
            PoseScript~\cite{posescript} & $\checkmark$ & $\times$ & $\times$ & $\checkmark$ & $\times$ & $\times$ \\
        PoseFix~\cite{delmas2023posefix} & $\checkmark$ & $\times$ & $\times$ & $\times$ & $\checkmark$ & $\checkmark$ \\
            Ours & $\checkmark$ & $\checkmark$ & $\checkmark$ & $\checkmark$ & $\checkmark$ & $\checkmark$ \\
            \bottomrule
        \end{tabular}
    }
    \vspace{-0.6cm}
    \label{tab:related_work}
\end{table}

\subsection{Discussion}
There are several related works worth discussing, which focus on linking textual descriptions to human poses. We have summarized
key differences in \cref{tab:related_work}: (1) PoseScript~\cite{posescript} utilizes generic rules to annotate textual descriptions from state-level 3D keypoints. In contrast, utilizing GPT-4V with well-designed prompts for annotation is more simple and fine-grained. Additionally, our annotation introduces state-to-state human and object transitions, as well as interactions, which PoseScript overlooks. (2) Although PoseFix~\cite{delmas2023posefix} annotates text descriptions of human state transitions, it similarly overlooks object state transitions and interaction transitions, which are critical and unique for the HOI problem.

\section{Fine-grained 3D HOI Tasks}
\label{sec:task}
\noindent \textbf{Motivation.}
Building on Semantic-HOI introduced in \cref{dataset_all}, we propose three novel tasks to showcase fine-grained semantic-aligned HOI, motivated by varying objectives: {\textbf{(1) Understanding:}} The task of captioning an HOI state (\ie, the poses of both the human and the object) with textual descriptions serve as the most straightforward method for demonstrating alignment. This approach addresses a gap often overlooked in existing tasks. {\textbf{(2) Reasoning:}} Building upon the understanding task, the depth of the problem can be extended to reason about the next HOI state via textual descriptions, when knowing the action goal.
{\textbf{(3) Generation:}} Moving forward, beyond simply considering the current or next state, comprehending the fine-grained movement descriptions to generate the next state based on the current one is promising.

\begin{figure*}[t]
    \centering
 	{
        \includegraphics[width=1.0\linewidth]{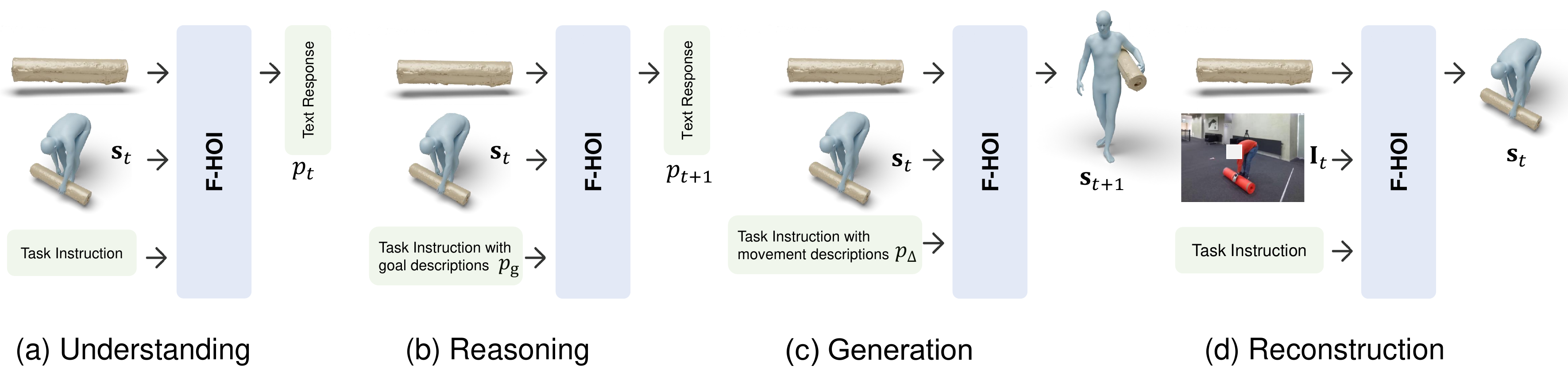}   
 	} 
  \vspace{-0.5cm}
\caption{Input and output definitions for each task.}
  \vspace{-0.5cm}
    \label{Fig:f_hoi_task}
\end{figure*}

\noindent \textbf{Problem Definitions.}
As shown in \cref{Fig:f_hoi_task}, conditioned by the object mesh and each task instruction, the proposed three tasks, along with the traditional reconstruction task, can be formulated as follows:
\textbf{(1) Understanding.} Given the $t$-th HOI state $\mathbf{s}_t$, the model aims to produce the corresponding fine-grained textual descriptions $p
_t$. Specifically, the HOI state $\mathbf{s}_t$ can be formulated as $(M(\theta,\beta), O)$. $M(\theta,\beta)$ is the human state obtained by a parametric human body model SMPL $M$~\cite{smpl}. $\theta$ and $\beta$ are pose and shape parameters, respectively. Following~\cite{feng2024chatpose}, $\beta$ is by default set to zeros, corresponding to the average body shape. $O$ is the object state represented by $6$ degrees of freedom ($6$DoF) object pose (\textit{i.e.}, translations and rotations).
\textbf{(2) Reasoning.} Given the goal descriptions $p_g$ (e.g., a person picks up a backup) and the $t$-th HOI state $\mathbf{s}_t$, we expect the model to reason the next state's text descriptions $p_{t+1}$.
\textbf{(3) Generation.} Given the $t$-th HOI state 
$\mathbf{s}_t$ and the movement descriptions $p
_\Delta$, the model needs to generate the next HOI state $\mathbf{s}_{t+1}$.
\textbf{(4) Reconstruction.} Given the 2D image $\mathbf{I}_t$ at $t$-th state, the model output the HOI state $\mathbf{s}_{t}$. However, directly outputting the object pose is challenging. We perform object-conditioned human reconstruction, which inputs the object pose at $t$-th state and outputs the human pose at $t$-th state.

\begin{figure*}[t]	
\centering
 	{
        \includegraphics[width=1.0\linewidth]{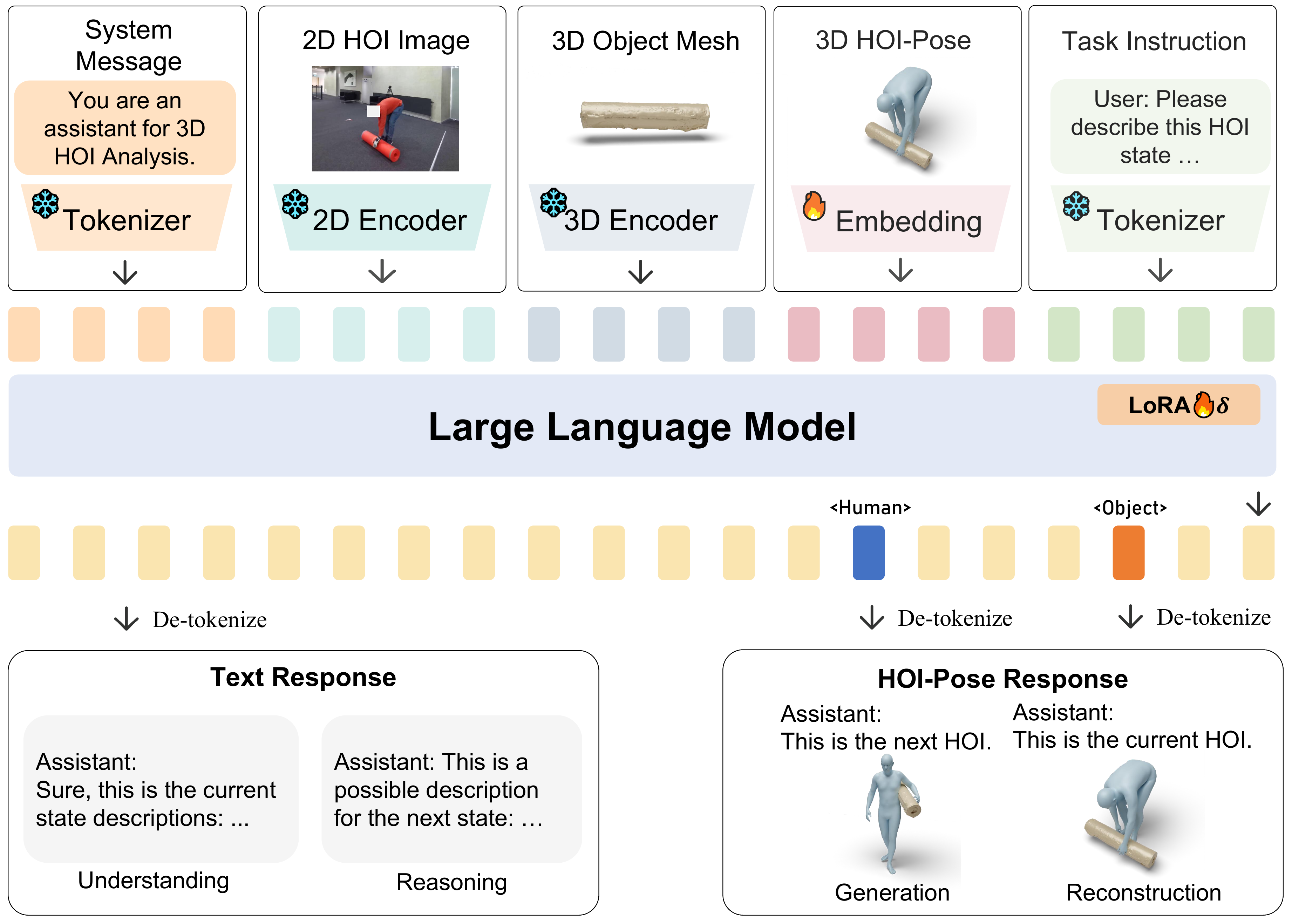}   
 	} 
  \vspace{-0.5cm}
    \caption{Overview of our \ModelName framework, which contains the three components: multi-modal encoders, a large language model, and task-specific projectors. Based on different task instructions, \ModelName could support multi-modal inputs and complete diverse HOI tasks, covering understanding, reasoning, generation, and reconstruction tasks.}
    \label{Fig:method}
            \vspace{-0.6cm}
\end{figure*}

\section{Model}

\subsection{Motivation}

As defined in \cref{sec:task}, the consistency across the four tasks essentially involves learning fine-grained alignments across 2D, 3D, and language spaces. Therefore, it is crucial to use a single model to unify the formulations of these four tasks and stimulate potential mutual benefits among them, which is demonstrated in our ablation study. On the other hand, due to the complexity of new tasks, previous technical paradigms~\cite{hassan2019resolving,chen2019holistic,weng2021holistic,xu2021d3d,siwei2021learning,zhang2022couch,wang2022humanise} cannot meet the requirements of tasks, which necessitate semantic comprehension and cognition capabilities for handling lengthy sentences in fine-grained descriptions. Based on the above motivations, we propose to empower the Multi-modal Large Language Model to complete the proposed HOI tasks with flexible inputs.

\subsection{Architecture}
As illustrated in \cref{Fig:method}, based on different task instructions, \ModelName could support multi-modal inputs and complete diverse HOI tasks containing the three components: multi-modal encoders, LLM Backbone, and task-specific projectors.

\noindent \textbf{Multi-modal Encoders.} 
We leverage different modality encoders to project each input modality into tokens that align together within the language space of the LLM backbone. Specifically,
(1) We utilize the SentencePiece tokenizer~\cite{kudo2018sentencepiece} to encode textual descriptions, including task instructions, goal descriptions $p_g$ in the reasoning task, and fine-grained movement descriptions $p_\Delta$ in the generation task, as in Fig.~\ref{Fig:f_hoi_task};
(2) We employ the frozen CLIP image encoder and one additional projection layer to encode 2D HOI images;
(3) We utilize the frozen 3D point encoder in Uni3D~\cite{zhou2023uni3d} with a trainable projection layer to encode 3D object meshes;
(4) For HOI-Pose, we use separate projection layers to project the human pose and object pose into the hidden space of the LLM backbone.

\noindent \textbf{LLM Backbone.} 
We utilize Vicuna-7B~\cite{zheng2023judging} as the LLM backbone and employ LoRA~\cite{hu2021lora}, which initializes a trainable matrix for fine-tuning the LLM.

\noindent \textbf{De-tokenlization.} (1) For text response, we also utilize the SentencePiece detokenizer to decode all the text tokens into textual descriptions; (2) For HOI-Pose response, we utilize a trainable human-pose projection layer to decode the special token \texttt{<Human>} into human pose, while employing another trainable object-pose projection layer to decode the special token \texttt{<Object>} into object pose.

\subsection{Training Pipeline}

\noindent \textbf{Pretraining for Alignments.} 
Thanks to the versatility of our model in handling various input and output formats, we are able to utilize multiple related datasets for pretraining, thereby improving alignments across different modalities. Specifically, acknowledging the critical role of human pose diversity in HOI tasks, we engage with extensive pose estimation and description datasets to facilitate alignment between images and human poses, as well as text and human poses. To this end, we employ the COCO~\cite{coco} and Posescript~\cite{posescript} datasets, respectively. We convert these two datasets into different question-answer formats and optimize the model using the following loss functions:
\begin{equation*}
     \mathcal{L}  =  \mathcal{L}_{\rm{text}}+\mathcal{L}_{\rm{pose}}
\end{equation*}
where $\mathcal{L}_{\rm{text}}$ is the cross-entropy loss typically applied 
for prefix language modeling such as GPT. $\mathcal{L}_{\rm{pose}} = \|\theta_{\rm{gt}} - \theta_{\rm{pred}} \|$ is the L1 loss computed between predicted human pose parameters and ground-truth ones.

\noindent \textbf{Multi-Task Instruction Tuning.} In this stage, we convert the proposed dataset into task-specific instruction-following format, including understanding, reasoning, generation, and reconstruction tasks. 
By joint training these tasks, we optimize the model using the following loss functions:
\begin{equation*}
     \mathcal{L}  =  \mathcal{L}_{\rm{text}}+\mathcal{L}_{\rm{hoi}}
\end{equation*}
where $\mathcal{L}_{\rm{hoi}} = \| \Delta\theta_{\rm{gt}} - \Delta\theta_{\rm{pred}} \| + \| \Delta O_{\rm{gt}} - \Delta O_{\rm{pred}} \|$ is the L1 loss to minimize the translation difference between predicted and ground-truth human and object pose parameters from the start state to the next state, which only applies to the generation (current state as start state) and reconstruction (default state as start state). 
Therefore, $\mathcal{L}_{\rm{hoi}}$ is computed as the offset. This approach could benefit the generation task, which we refer to as \textbf{offset regression}.

\section{Experiment}

\subsection{Experimental Setup}

\noindent \textbf{Evaluation Metric.} Since the output of understanding and reasoning tasks are textual descriptions, we employ two commonly used metrics from the NLP field for evaluations: (1) BLUE-4~\cite{papineni2002bleu} analyzes the co-occurrences of $4$-grams between the
predicted and ground-truth sentences.
(2) ROUGH~\cite{lin2004rouge} examines the adequacy and fidelity of the predicted sentences based on recall rate. 
In contrast, for evaluating generation and reconstruction tasks, we follow previous works~\cite{xie2022chore,zhang2020perceiving,bhatnagar2022behave} to utilize the Chamfer distance for both humans and objects. 
Specifically, we decouple the human pose into different body parts for evaluation, including the head, two arms, two hands, and two legs, to better demonstrate the performance guided by fine-grained semantic descriptions.

\noindent \textbf{Baseline.} 
Since this work focuses on entirely new HOI tasks that existing models have not been able to tackle, we choose the base multi-modal large language model in \ModelName for comparison. Specifically, we use our \DataName to finetune LLaVA-1.5V-7B~\cite{llava} as our baseline model, which incorporates Vicuna-7B~\cite{zheng2023judging} as the LLM backbone with a CLIP image encoder~\cite{clip} for visual encoding. Considering the original LLaVA only takes images and textual descriptions as inputs, we further incorporate 3D HOI-Pose embedding to process HOI poses, enabling the completion of our tasks. In contrast, we do not input the object mesh and instead output the HOI-Pose as a textual response. To obtain a complete and precise HOI-Pose, we decouple the human pose parameters and the object pose, and then perform multi-turn conversation for training and the batch inference to query them separately.

\noindent \textbf{Implementation Details.}
We employ LLaVA-1.5V-7B to initialize the model weight. During training, we freeze the CLIP image encoder while fine-tuning the LLM using LoRA~\cite{hu2021lora}. Additionally, we train projection layers to adapt the LLM to our proposed HOI tasks. For fine-tuning with LoRA, we set the rank to $128$ and the alpha to $256$. We employ AdamW~\cite{loshchilov2017decoupled} for network optimization, with a learning rate of $2e-4$ and weight decay of $0$.
During training, we utilize $8$ Nvidia A100 GPUs, each with 80G of memory. We set the batch size to $16$ for each GPU and configure a gradient accumulation step of $1$.

\subsection{Main Results}
\label{sec:q1}

\begin{figure*}[h]
    \centering
 	{
        \includegraphics[width=1\linewidth]{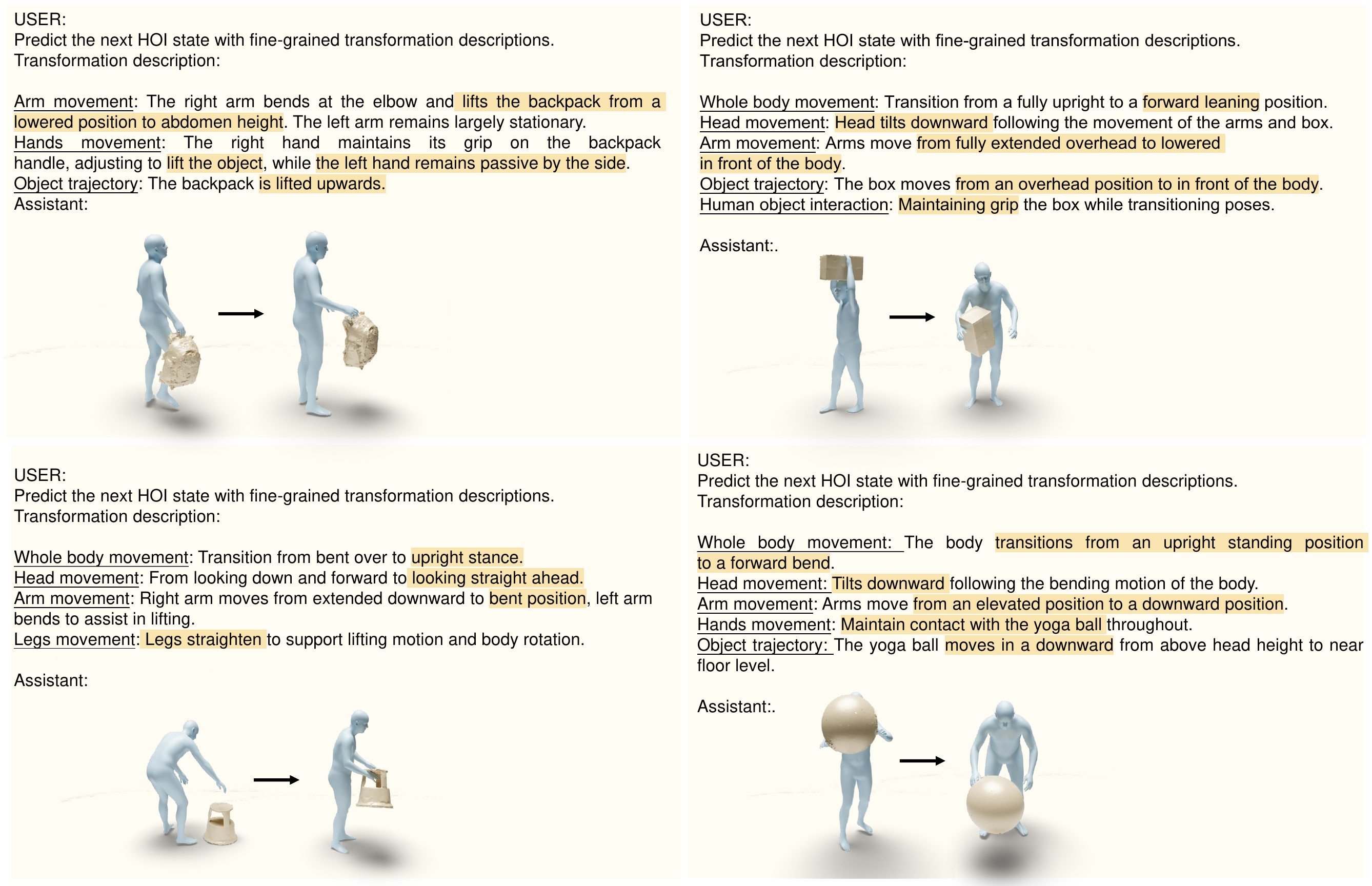}   
 	} 
  \vspace{-0.5cm}
\caption{Qualitative results of \ModelName on generation task.}
    \label{Fig:vis_generation}
            \vspace{-0.2cm}
\end{figure*}

For quantitative results, as illustrated in \cref{tab:understanding}, ~\cref{reasoning}, ~\cref{tab:generation}, and ~\cref{tab:reconstruction}, we conduct a comprehensive comparative analysis of our proposed model, \ModelName, across all four tasks against the established baseline, utilizing our \DataName dataset. \ModelName significantly and consistently outperforms its baseline, demonstrating its effectiveness in achieving fine-grained semantic alignment. Moreover, we provide the qualitative results as shown in \cref{Fig:vis_generation} for the generation task.
Overall, \ModelName can learn rich semantic representations at the state level,
adeptly tackling our proposed understanding, reasoning, and generation tasks.

\begin{table*}[h]
\vspace{-0.5cm}
    \begin{center}
                            \begin{minipage}{0.4\linewidth}
                            	\caption{Understanding Task.}
                             \vspace{-0.3cm}
                            \centering
        \resizebox{\linewidth}{!}{
            \makeatletter\def\@captype{table}\makeatother
		\begin{tabular}{l|cccc}
		\toprule
		\textbf{Method}  & BLEU-4$\uparrow$~\cite{papineni2002bleu}  & ROUGE$\uparrow$~\cite{lin2004rouge}  \\
		\midrule
     Baseline & 20.09  & 35.20 \\
    \ModelName & \textbf{26.78}  &  \textbf{45.29} \\ 
		\bottomrule
		\end{tabular}%
            }
\label{tab:understanding}
        \end{minipage}
            \hfill
                \begin{minipage}{0.4\linewidth}
                \caption{Reasoning Task.}
                \vspace{-0.3cm}
        \resizebox{\linewidth}{!}{
            \makeatletter\def\@captype{table}\makeatother
		\begin{tabular}{l|cccc}
		\toprule
		\textbf{Method}  & BLEU-4$\uparrow$~\cite{papineni2002bleu}  & ROUGE$\uparrow$~\cite{lin2004rouge}  \\
		\midrule
     Baseline  & 19.51 & 35.69\\
    \ModelName & \textbf{25.56}  & \textbf{41.84}\\ 
		\bottomrule
		\end{tabular}%
            }
    \label{reasoning}
        \end{minipage}
 \vspace{-0.5cm}
    \end{center}

\end{table*}

\begin{table*}[h]
	\centering
 		\caption{
			{Generation Task.}
}
\vspace{-0.3cm}
	\resizebox{1\linewidth}{!}
	{
	\setlength{\tabcolsep}{1mm}
		\begin{tabular}{l|cccccccc|c}
		\toprule
		\textbf{Method} & Head & Left Arm  & Right Arm &Left Hand &  Right Hand & Left leg & Right Leg  & Object      & {Averaged}$\downarrow$ \\
		\midrule 
     Baseline & 36.8 & 28.1 &34.0 & 55.1& 73.7 & 23.4 & 30.9 &  72.9 & 44.4 \\

    \ModelName & \textbf{16.8} & \textbf{13.5} & \textbf{16.2} & \textbf{35.2} & \textbf{42.3} & \textbf{10.6}&\textbf{14.1}& \textbf{34.8} &  \textbf{22.9}  \\
		\bottomrule
		\end{tabular}%
		}
\label{tab:generation}
 \vspace{-0.3cm}
\end{table*}

\begin{table*}[h]
	\centering
 		\caption{
			Object-conditioned Reconstruction Task.
}
\vspace{-0.3cm}
	\resizebox{1\linewidth}{!}
	{
	\setlength{\tabcolsep}{1mm}
		\begin{tabular}{l|ccccccc|c}
		\toprule
		\textbf{Method} & Head & Left Arm  & Right Arm &Left Hand &  Right Hand & Left leg & Right Leg       & {Averaged}$\downarrow$ \\
		\midrule 
     Baseline  & 45.8 & 37.3 & 42.8 & 62.5 & 79.4 & 32.9 & 39.4 & 48.6 \\
    \ModelName &  \textbf{20.8} & \textbf{15.1} & \textbf{19.5} & \textbf{38.7} & \textbf{50.4} & \textbf{11.6} & \textbf{16.9} & \textbf{24.7} \\
		\bottomrule
		\end{tabular}%
		}
\vspace{-0.5cm}
\label{tab:reconstruction}
\end{table*}

\subsection{Ablation Study}

\noindent \textbf{Offset Regression.} For the generation task, \ModelName designs an offset-based regression method, which expects the model to predict offsets from the input HOI-Pose, thereby supervising the HOI-Pose offsets to achieve better alignment with movement descriptions. The results in \cref{tab:ablation} indicate that this regression approach can notably enhance the generation task by achieving improved alignment with movement descriptions.

\begin{table*}[h]
\vspace{-0.6cm}
		\caption{
			Effect of offset regression on different HOI tasks. We adopt the BLUE-4 to evaluate understanding and reasoning tasks and use the averaged Chamfer distance for generation and reconstruction tasks.
}
\vspace{-0.3cm}
	\centering
	\resizebox{0.8\linewidth}{!}
	{
	\setlength{\tabcolsep}{1mm}
		\begin{tabular}{c|cccc}
		\toprule
	 Offset & Understanding$\uparrow$ & Reasoning$\uparrow$    & Generation$\downarrow$  & Reconstruction$\downarrow$ \\
		\midrule
            \ding{55} & \textbf{26.91} &  24.73 & 27.9 & 25.5\\
                   \checkmark & 26.78   & \textbf{25.56 }  & \textbf{22.9}  & \textbf{24.7}  \\
		\bottomrule
		\end{tabular}%
		}
\vspace{-0.6cm}
\label{tab:ablation}
\end{table*}

\begin{table*}[t]
	\centering
 		\caption{
			Effect of image-to-pose and text-to-pose alignment on different HOI tasks.
}
  \vspace{-0.3cm}
	\resizebox{1\linewidth}{!}
	{
	\setlength{\tabcolsep}{1mm}
		\begin{tabular}{c|cc|cccc}
		\toprule
	No.&	Image-to-Pose & Text-to-Pose  & Understanding$\uparrow$ & Reasoning$\uparrow$    & Generation$\downarrow$  & Reconstruction$\downarrow$ \\
		\midrule
   \#1 &   \ding{55} &  \ding{55}  & 21.21 & 20.98 & 32.7  & 33.9  \\
   \#2 &   \checkmark  &  \ding{55} & 23.76 & 23.43 &  24.8 &  25.8 \\
     \#3 &  \checkmark &  \checkmark & \textbf{26.78}   & \textbf{25.56 }  & \textbf{22.9}  & \textbf{24.7} \\
		\bottomrule
		\end{tabular}%
		}
    \vspace{-0.5cm}
\label{tab:ablation_pretraining}
\end{table*}

\begin{table*}[t]
	\centering
 		\caption{
			Effect of joint training across multiple tasks on each HOI task.
}
\vspace{-0.3cm}
	\resizebox{0.8\linewidth}{!}
	{
	\setlength{\tabcolsep}{1mm}
		\begin{tabular}{l|cccc}
		\toprule
		\textbf{Task}  &  Understanding$\uparrow$ & Reasoning$\uparrow$    & Generation$\downarrow$  & Reconstruction$\downarrow$ \\
		\midrule
     Understanding&  24.67 & - & - & -\\
          Reasoning& -&23.89 & -& - \\
                    Generation& -& -& 24.0 & - \\
                        Reconstruction& -& -& -& 26.8  \\
    All &  \textbf{26.78}   & \textbf{25.56 }  & \textbf{22.9}  & \textbf{24.7}  \\
		\bottomrule
		\end{tabular}%
		}

\label{tab:ablation_task}
\vspace{-0.3cm}
\end{table*}

\begin{figure*}[h]
    \centering
 	{
        \includegraphics[width=1\linewidth]{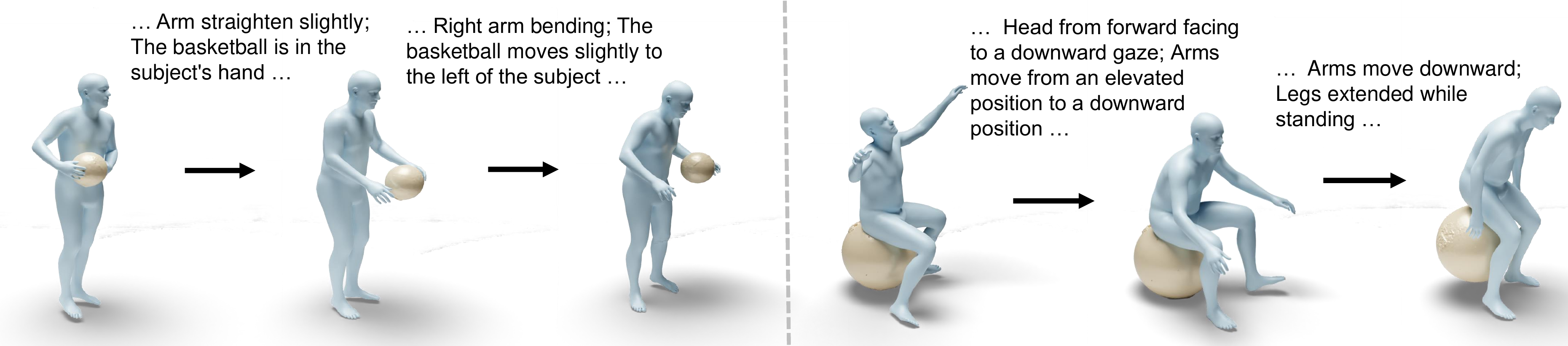}   
 	} 
  \vspace{-0.5cm}
\caption{We show that \ModelName has the potential to utilize fine-grained descriptions at the state level for performing sequence generation state-by-state.}
    \label{Fig:Fi}
      \vspace{-0.6cm}
\end{figure*}

\noindent \textbf{Image-to-Pose Alignment.} Thanks to the input and output flexibility of our model, we can use large-scale image-pose paired dataset COCO~\cite{coco} to perform image-to-pose alignment. The results in \cref{tab:ablation_pretraining}, \#1 vs. \#2, demonstrate significant improvements in both the reconstruction and generation tasks. The enhanced human pose diversity is crucial for effective HOI, thereby contributing to the observed improvements.

\noindent \textbf{Text-to-Pose Alignment.} 
We utilize the text-pose paired dataset PoseScript~\cite{posescript} to achieve text-to-pose alignment. The results in \cref{tab:ablation_pretraining}, comparing methods \#2 and \#3, demonstrate notable benefits for understanding and reasoning tasks. This approach effectively aligns poses with diverse descriptions, contributing to improved performance.

\noindent \textbf{Multi-task Joint Training.} In the previous ablation studies, a consistent phenomenon is observed: an improvement in one task's performance often leads to improvements in other tasks as well. Furthermore, we conduct ablations focusing on single tasks. The results in \cref{tab:ablation_task} demonstrate the presence of mutual benefits across different tasks in multi-task training, significantly outperforming single-task training. In addition, training with multiple modalities input of the same sample (e.g., images, HOI-Pose, and textual descriptions) can also provide more information to improve performance.

\section{Discussion}
 \noindent \textbf{State-by-state Generation for an HOI process.} Although the primary focus of this work is to align fine-grained semantic details with HOI at the state level, we demonstrate the potential of this paradigm to generate long sequences for the interaction process while maintaining a detailed understanding of each intermediate state and the transitions between states, as shown in Fig~\ref{Fig:Fi}.

\begin{figure}[t]
    \centering
    \includegraphics[width=\linewidth]{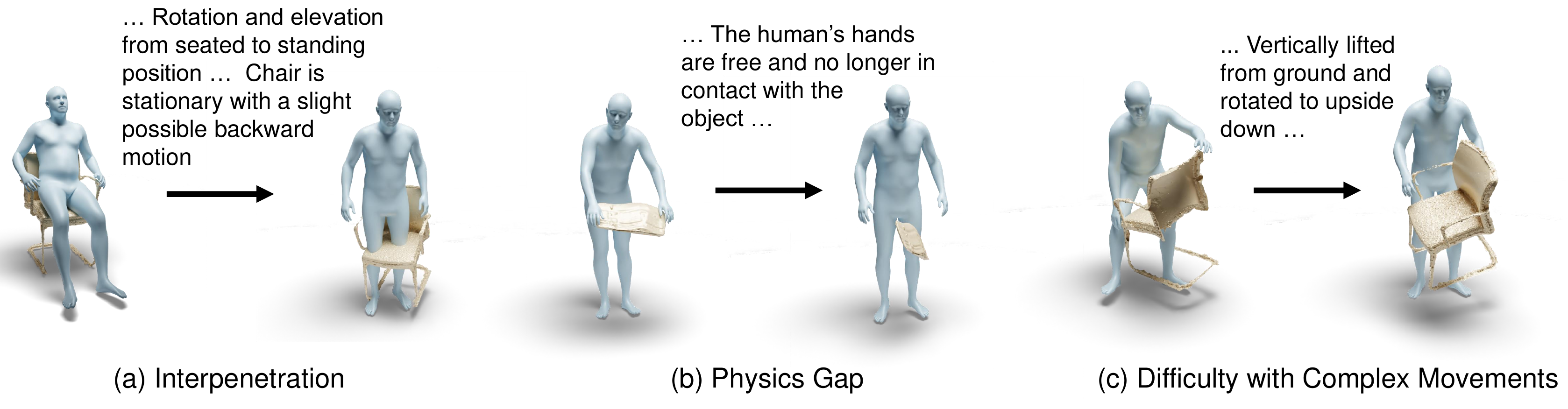}
     \vspace{-0.5cm}
    \caption{We show three types of failure cases in our method.}
    \label{fig:failure}
    \vspace{-0.5cm}
\end{figure}

\noindent \textbf{Failure Case Analysis.}
We show three types of failure cases in our method, as shown in \cref{fig:failure}. \textbf{(1) Interpenetration:} \ModelName employs fine-grained textual supervision (e.g., body part with contact) for implicit human-object optimization. Compared with previous methods~\cite{hassan2019resolving,chen2019holistic,weng2021holistic,xu2021d3d,siwei2021learning,zhang2022couch,wang2022humanise,wang2022reconstructing,xu2021d3d}, it is conceptually simple by eliminating complex structured HOI modeling and explicit contact supervision~\cite{diller2023cg,xiao2023unified}. However, in cases where there are conflicts between human actions and objects, the interpenetration still persists, as illustrated in \cref{fig:failure}-(a).
\textbf{(2) Physics Gap:} 
Since \ModelName does not explicitly incorporate the physical laws~\cite{xu2023interdiff,xiao2023unified,wang2023physhoi,xie2021physics,yuan2023physdiff}, the generation of the next HOI state merely aligns with linguistic descriptions without constraints imposed by physical reality. For instance, as depicted in \cref{fig:failure}-(b), without contact, the ``keyboard'' would actually fall to the ground due to gravity. 
\textbf{(3) Difficulty with Complex Movements:} \ModelName fails to generate the next HOI state when the trajectory of object states between the current and next states is complex or highly variable, as exemplified in \cref{fig:failure}-(c). This is primarily due to our dataset lacking sufficient examples of long-term and complex movements. One direct solution to address this issue is to decompose the description of movements into multiple fine-grained stages.

\noindent \textbf{Limitations.} Overall, our work serves as a pioneering work that provides the community with a new perspective for fine-grained semantic-aligned 3D human-object interaction modeling. However, as an early-stage effort, our work leaves ample room for further exploration in this field. Here, we discuss several limitations to inspire future research:
\textbf{(1)} As shown in \cref{Fig:f_hoi_task}, our proposed tasks require inputs including a 3D object mesh, 3D HOI-Pose, and textual descriptions. These input requirements significantly hurt the convenience of inference. Reducing the strict requirements of input modalities is an area worth exploring.
\textbf{(2)} Our work only evaluates the effectiveness of fine-grained textual descriptions and HOI-Pose alignment in closed-set scenarios within existing datasets. However, such a model lacks generalization to open-set scenarios, which is limited by interaction diversity and unseen object meshes.
\textbf{(3)} Based on our task and problem definitions, \ModelName can only perform long sequence generation state-by-state through fine-grained textual descriptions for addressing the HOI motion generation task. This approach introduces complexity and error accumulation, and it does not address the issue of smooth transitions between states.
\textbf{(4)} The design of \ModelName follows a simple and intuitive principle and could be considered as a baseline model.
It may perform poorly compared to previous methods~\cite{zhang2022couch,zhang2020perceiving,xie2022chore,diller2023cg} in generation and reconstruction tasks.
\textbf{(5)} 
We expect our model to predict hand parameters to better demonstrate the alignment between fine-grained textual descriptions and HOIs. However, our preliminary results indicate that our model underperforms in capturing the details of the hands.
\noindent \textbf{(6)} For understanding and reasoning tasks, \ModelName heavily relies on the priors of large language models. Despite showing significant potential, we still identify several understanding and reasoning errors, such as inaccurate judgments of interactions and incorrect assessments of the spatial relationships between body parts. These limitations arise from the restricted data volume used to align HOI-Pose with fine-grained descriptions, while the richness and quality of the textual descriptions also affect performance~\cite{chen2023sharegpt4v}.

\noindent \textbf{Future Directions.} 
\noindent \textbf{(1) Flexiable Input Modality.} As discussed in the limitations above, reducing the required input modalities is worth considering. For instance, the object mesh could potentially be obtained through a text-to-3D approach~\cite {poole2022dreamfusion,chen2024text,lin2023magic3d}. Furthermore, the initial HOI-Pose could be directly derived from image input, as the human SMPL parameters and object 6DoF pose can be obtained by other powerful models~\cite{lin2023one,wen2024foundationpose,corsetti2024open,lin2024sam,yang2023unipose,yang2023neural,bogo2016keep}. \textbf{(2) Increase Data Scale.} Our \DataName currently covers only three datasets with a limited number of samples. Merging more HOI datasets~\cite{liu2022hoi4d,zhang2024force}, scaling up the number of samples, and enriching the textual descriptions are also worth exploring. Moreover, exploring the hierarchy of human body part states, object states, and actions are promising and meaningful~\cite{belkhale2024rt}.  \textbf{(3) Diverse Model Architectures.} Due to the complexity of new tasks and the limited data samples, our model is built on the Multi-modal Large Language Model, which brings semantic comprehension and cognitive capabilities for handling lengthy sentences in fine-grained descriptions. However, compared to previous HOI models, our model has a significantly larger amount of parameters, making it less lightweight for addressing HOI tasks. Thus, as data scales up, exploring other model architectures~\cite{tevet2022motionclip} also becomes an important consideration.

\section{Conclusion}
This paper proposes the overlooked challenge of fine-grained semantic-aligned 3D human-object interaction (HOI), which is inadequately addressed by current HOI datasets and models. To bridge this gap, we introduce Semantic-HOI, a new dataset featuring over 20K meticulously annotated HOI state pairs, each equipped with detailed descriptions and corresponding body movements between consecutive states. Leveraging this dataset, we formulate three state-level HOI tasks aimed at achieving fine-grained semantic alignment within the HOI sequence.
Moreover, we present \ModelName, which empowers the MLLM to proficiently tackle the proposed HOI tasks. Extensive experiments showcase \ModelName's prowess in aligning HOI states with fine-grained semantic descriptions.

\section*{Acknowledgement}
The work is partially supported by the Young Scientists Fund of the National Natural Science Foundation of China under grant No.62106154, by the Natural Science Foundation of Guangdong Province, China (General Program) under grant No.2022A1515011524, and by Shenzhen Science and Technology Program JCYJ20220818103001002, and by the Guangdong Provincial Key Laboratory of Big Data Computing, The Chinese University of Hong Kong (Shenzhen).

\bibliographystyle{splncs04}
\bibliography{main}
\end{document}